\documentclass[10pt,twocolumn,letterpaper]{article}

\usepackage{wacv}
\usepackage{times}
\usepackage{epsfig}
\usepackage{graphicx}
\usepackage{amsmath}
\usepackage{amssymb}
\usepackage{booktabs}
\usepackage{multirow}
\usepackage{bbm}

\wacvfinalcopy % *** Uncomment this line for the final submission

\ifwacvfinal
\usepackage[breaklinks=true,bookmarks=false]{hyperref}
\else
\usepackage[pagebackref=true,breaklinks=true,colorlinks,bookmarks=false]{hyperref}
\fi

\begin{document}

%%%%%%%%% TITLE
\title{TAX: Tendency-and-Assignment Explainer
\\for Semantic Segmentation with Multi-Annotators}

\author{
Yuan-Chia Cheng\\
National Taiwan University\\
{\tt\small r08942154@ntu.edu.tw}\\
\and
Zu-Yun Shiau\\
National Taiwan University\\
{\tt\small r09942069@ntu.edu.tw}\\
\and
Fu-En Yang\\
National Taiwan University\\
{\tt\small f07942077@ntu.edu.tw}\\
\and
Yu-Chiang Frank Wang\\
National Taiwan University, NVIDIA\\
{\tt\small ycwang@ntu.edu.tw}\\
}

\maketitle
\thispagestyle{empty}

%%%%%%%%% ABSTRACT
\begin{abstract}
To understand how deep neural networks perform classification predictions, recent research attention has been focusing on developing techniques to offer desirable explanations. However, most existing methods cannot be easily applied for semantic segmentation; moreover, they are not designed to offer interpretability under the multi-annotator setting. Instead of viewing ground-truth pixel-level labels annotated by a single annotator with consistent labeling tendency, we aim at providing interpretable semantic segmentation and answer two critical yet practical questions: ``who" contributes to the resulting segmentation, and ``why" such an assignment is determined. 

In this paper, we present a learning framework of Tendency-and-Assignment Explainer (TAX), designed to offer interpretability at the annotator and assignment levels. More specifically, we learn convolution kernel subsets for modeling labeling tendencies of each type of annotation, while a prototype bank is jointly observed to offer visual guidance for learning the above kernels. For evaluation, we consider both synthetic and real-world datasets with multi-annotators. We show that our TAX can be applied to state-of-the-art network architectures with comparable performances, while segmentation interpretability at both levels can be offered accordingly.
\end{abstract}

%%%%%%%%% BODY TEXT
\section{Introduction}
\label{sec:intro}

\begin{figure}[t!]
	\centering
	\includegraphics[width=0.45\textwidth]{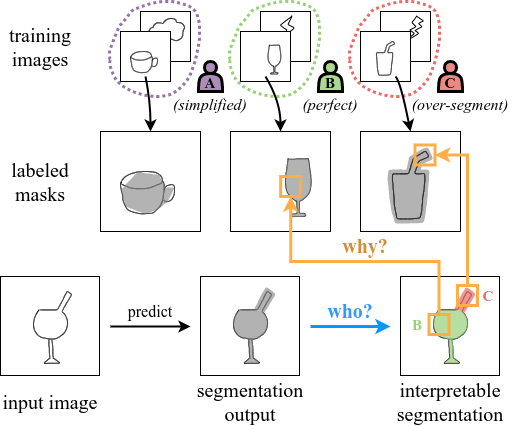}
 	\vspace{0mm}
    \caption{\textbf{Illustration of interpretable semantic segmentation with training data annotated by multi-annotators.} Practically, each annotator has his/her labeling tendency. We aim to address semantic segmentation, while providing interpretability at both annotator and assignment levels achieved (denoted in blue and orange arrows, respectively).}
    \vspace{0mm}
	\label{fig:teaser}
\end{figure}

Deep learning has presented its remarkable capability spanning a wide variety of computer vision tasks, such as classification, segmentation, and object detection. Despite the remarkable performances, deep learning models are often viewed as unintelligible black boxes as the developers or the users are not able to tell why and how such models function. In real-world applications, the explainability of deep learning models and their prediction outputs are with increasing demands. To this end, researchers begin to shift their efforts to develop interpretable deep learning models.

Recent interpretable deep learning in computer vision has made notable progress, particularly for classification models. A plethora of approaches have been proposed to derive map-based~\cite{dabkowski2017real, selvaraju2017grad, smilkov2017smoothgrad, xu2020attribution, kapishnikov2019xrai, fong2017interpretable, sundararajan2017axiomatic, chattopadhay2018grad, zhou2016learning, huang2020interpretable} and example-based~\cite{chen2018learning, kanehira2019learning, jeyakumar2020how, gulshad2020explaining, chen2019looks, rymarczyk2020protopshare} explanation, which points out how local information is engaged in the image-level classification process. As for interpretable segmentation, methods of \cite{mickstrom2018uncertainty, sun2020saunet} can be viewed as map-based manners built upon the backpropagation techniques or attention mechanisms. 
However, most existing works towards interpretable learning models cannot easily offer explanations to semantic segmentation with training image data annotated by multiple and possibly diverse annotators.

In real-world applications, one typically requires multiple experts for data labeling, since it would be time-consuming for one expert to annotate the entirity of a dataset. Ideally, all the annotators are supposed to abide by a standard annotating protocol or guideline (\eg, the PASCAL VOC labeling guideline\footnote[1]{http://host.robots.ox.ac.uk/pascal/VOC/voc2012/guidelines.html}). In practice, each annotator holds different labeling tendencies due to their backgrounds. As noted in~\cite{zhang2020disentangling}, some tend to over-segment image boundaries while some might under-segment such data.
Taking the multi-annotator context into consideration, \cite{zhang2020disentangling} proposes a coupled model comprised of an annotator network estimating annotator confusion matrices when predicting the segmentation output. However, existing methods like~\cite{zhang2020disentangling, baumgartner2019phiseg} are not designed to offer any type of interpretability during inference.

Instead of treating the label quality or labeling tendency equally the same for each annotator, we aim at designing segmentation models with annotation interpretability. To produce proper interpretations for multi-annotator semantic segmentation as depicted in Figure~\ref{fig:teaser}, the segmentation model is expected to answer two questions of interest: ``who" contributes to the resulting segmentation tendency (\ie, annotator-level explanation), and ``why" the model determines the corresponding tendency during inference (\ie, assignment-level explanation). To realize the goal, we present a novel learning framework of \textit{Tendency-and-Assignment Explainer (TAX)}, which learns to perform interpretable segmentation, which jointly learns annotator-specific masks and the corresponding convolution kernel subsets. More specifically, the above kernel subsets are learned to describe the labeling tendencies of each annotator. Moreover, with the annotator-dependent prototype bank, our model allows to explain why particular annotator labeling tendency is utilized during inference, realizing explainable and traceable segmentation outputs. As confirmed by our experiments, our TAX can be applied to existing segmentation models, producing satisfactory and explainable segmentation outputs at annotator and assignment levels.

The contributions of this paper are highlighted below:
\begin{itemize}
\item To the best of our knowledge, we are among the first to offer interpretability for semantic segmentation models with the \textit{multi-annotator} setting.
\item Instead of assuming ground-truth labels are with the same annotation style, our TAX learning framework is designed to model specific labeling tendency, aiming at providing both annotator and assignment-level explanations for segmentation.
\item To offer interpretability at the annotator level, our TAX is able to learn distinct convolutional kernel subsets given training data annotated by different annotators.
\item Our TAX learns a prototype bank for describing image data, serving as explainable guidance for the assignment and learning of the above kernel subsets.
\end{itemize}

\section{Related Works}
\label{sec:related_works}

\subsection{Semantic Segmentation}
A great number of deep learning approaches have been proposed for semantic segmentation. As the pioneers, FCN~\cite{long2015fully} and U-Net~\cite{ronneberger2015u} have demonstrated excellent segmentation ability by adopting encoder-decoder architectures. To better capture information at multiple resolutions/levels, PSPNet~\cite{zhao2017pyramid} incorporates spatial pyramid pooling~\cite{lazebnik2006beyond} at different grid scales, whereas DeepLabv3~\cite{chen2017rethinking} utilizes Atrous Spatial Pyramid Pooling at different dilated convolution rates. As the final extension of DeepLab models~\cite{chen2015semantic, chen2017deeplab, chen2017rethinking}, DeepLabv3+~\cite{chen2018encoder} further exploits multi-scale contextual information by designing a simplified decoder module which refines the segmentation results along with object boundaries. On the other hand, EfficientPS~\cite{mohan2021efficientps} proposes a semantic aggregation head with a Mask RCNN-based~\cite{he2017mask} head to encode and fuse semantically rich multi-scale features.

In spite of impressive performances, existing models require collection of a large number of ground-truth annotated data, which are often provided by multiple users/experts. Expecting diverse segmentation tendencies, this leads to the challenging \textit{multi-annotator semantic segmentation} problem. Recently, Probabilistic U-Net~\cite{kohl2018probabilistic} and PHiSeg~\cite{baumgartner2019phiseg} exploit probabilistic CNN models to model the inter-reader variations in segmentation labels. And \cite{zhang2020disentangling} employs an annotator network which estimates individual confusion matrices facilitating the segmentation network to derive unobserved true label distribution. 
Nevertheless, existing works generally focus on learning reliable information across annotators, instead of providing outputs with proper explanation (\eg, interpretable and traceable segmentation outputs). To the best of our knowledge, no existing works can be directly applied to offer interpretability under this multi-annotator semantic segmentation setting.

\begin{figure*}[t!]
	\centering
	\includegraphics[width=0.99\textwidth]{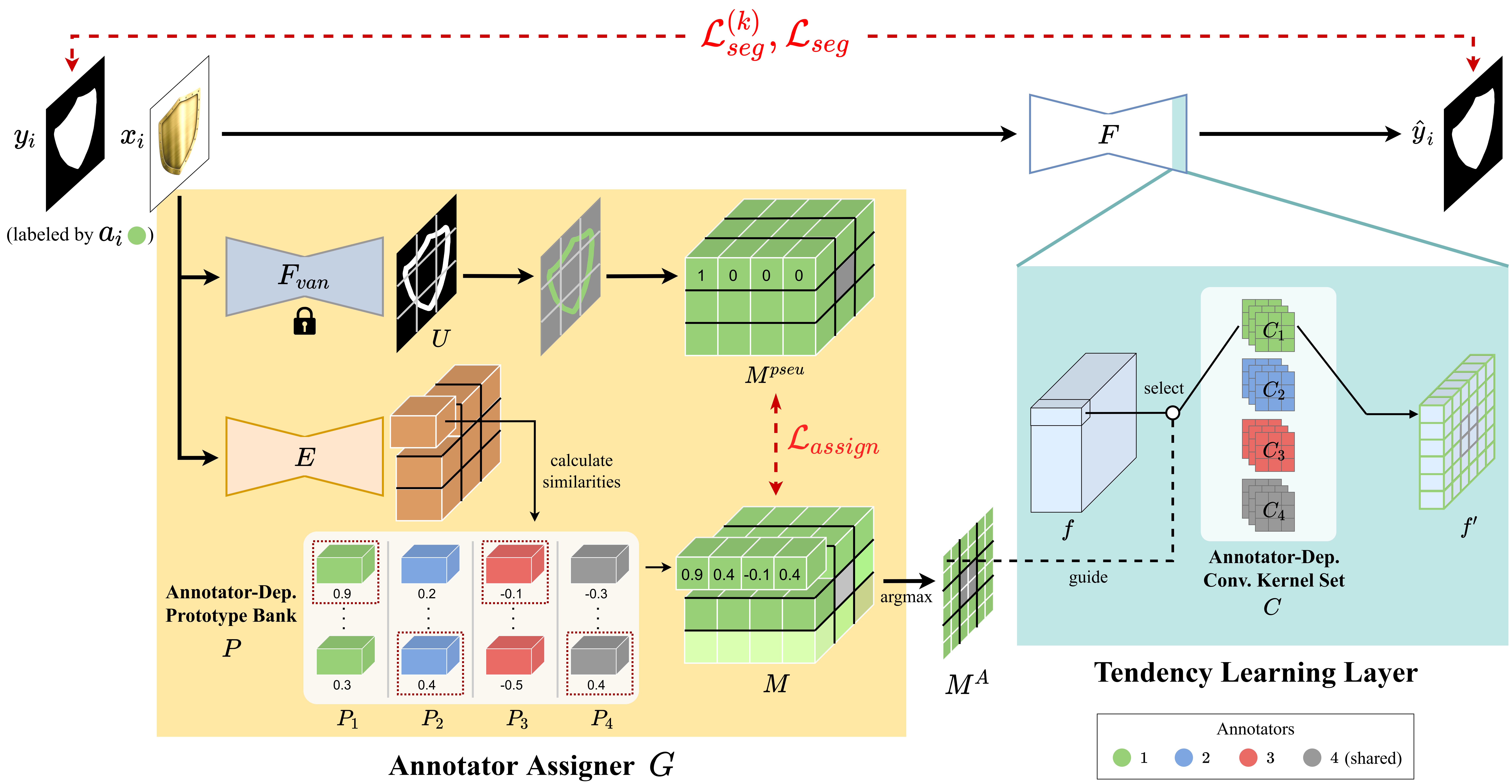}
 	\vspace{0mm}
    \caption{\textbf{Overview of our Tendency-and-Assignment Explainer (TAX) learning framework. } Given an input image $x_i$ annotated by $a_i$, the annotator assigner $G$ predicts an annotator mask $M^{A}$ via the learned annotator-dependent prototype bank $P$, guiding the segmentation model $F$ to learn annotator-dependent convolution kernel set $C$ in its tendency learning layer for providing explainable segmentation outputs. Best viewed in color.}
    \vspace{0mm}
	\label{fig:overview}
\end{figure*}

\subsection{Interpretable Deep Models}
With the need of understanding the decision process of deep learning models, a number of approaches have been developed. For image classification, the output interpretability majorly falls into two types: map-based and example-based. Map-based approaches~\cite{dabkowski2017real, selvaraju2017grad, smilkov2017smoothgrad, xu2020attribution, kapishnikov2019xrai, fong2017interpretable, sundararajan2017axiomatic, chattopadhay2018grad, zhou2016learning, huang2020interpretable} offer visualization maps to highlight the local parts of the input which contribute to the classification result most. Specifically, some~\cite{selvaraju2017grad, smilkov2017smoothgrad, xu2020attribution, kapishnikov2019xrai, fong2017interpretable, sundararajan2017axiomatic, chattopadhay2018grad} estimate saliency maps by disturbing the input elements or backpropagating the prediction to the input space, whereas some~\cite{zhou2016learning, huang2020interpretable} output attention maps by incorporating CAM or attention mechanism into models. As for the example-based branch~\cite{chen2018learning, kanehira2019learning, jeyakumar2020how, gulshad2020explaining, chen2019looks, rymarczyk2020protopshare}, one generally explains the prediction with instance-wise examples in place of pixel-wise importance weights. For instance, \cite{chen2019looks, rymarczyk2020protopshare} extract representative and discriminative part-wise prototypes which are then taken as the references for conditional prediction.

Recent works like~\cite{mickstrom2018uncertainty, sun2020saunet} extend the above map-based techniques for interpretable semantic segmentation. \cite{mickstrom2018uncertainty} applies guided backpropagation~\cite{springenberg2014striving} to FCN~\cite{long2015fully} for deriving attention maps, focusing on the object boundaries. On the other hand, mitigating the time-inefficient post hoc overhead like backpropagation, \cite{sun2020saunet} explicitly fuses attentive modules into texture and shape streams, where the texture stream handles coarse information for spatial attention maps while the shape stream is in charge of fine boundary information for shape attention maps. Nevertheless, the above methods are not designed for multi-annotator semantic segmentation scenarios, and thus the associated interpretability cannot be easily achieved.

\section{Proposed Method}
\label{sec:proposed_method}

\subsection{Problem Formulation and Method Overview}
We first define the task of \textit{interpretable multi-annotator semantic segmentation}, including the notations used in this paper. In the multi-annotator scenario, we have the $i$-th triplet as $(x_i, y_i, a_i)\in \{X, Y, A\}$ for training purposes. With a total of $N$ annotators available, each image $x_i$ is labeled by the corresponding annotator $a_i = k$ ($k$ as the annotator index/label), providing the associated ground-truth segmentation mask $y_i$. Note that the resolutions of $x_i$ and $y_i$ are both of size $H\times W$ pixels.

In this paper, we propose a learning framework of \textit{Tendency-and-Assignment Explainer (TAX)}, consisting of a \textit{tendency-preserving segmentation model} $F$ and an \textit{annotator assigner} $G$ (see Figure~\ref{fig:overview}). We deploy and learn a set of \textit{annotator-dependent convolution kernels} $C$ in $F$, aiming to describe the labeling tendency of each individual. On the other hand, annotator assigner $G$ is to produce annotator masks $M^{A}$ which guide the learning of the above kernels $C$. During inference, the annotator assigner $G$ predicts the annotator mask $M^{A}$ for the input image, with $F$ inferring the segmentation prediction using convolution kernels of the assigned annotators. With the above inference process, interpretability at annotator and assignment levels can be simultaneously achieved.

\subsection{Learning to Describe Labeling Tendencies}
\label{subsec:learn_to_represent}
To acquire tendency-explainable ability, we need the segmentation model to capture individual labeling tendencies from multi-annotated training data. To this end, we introduce a tendency-preserving segmentation model $F$ in Figure~\ref{fig:overview}, which utilizes existing segmentation models (\eg, U-Net~\cite{ronneberger2015u} or DeepLabv3+~\cite{chen2018encoder}) with only the last $3\times 3$ convolution layers modified.

In particular, we propose to learn the \textit{annotator-dependent convolution kernel set} $C$ in these final layers, since labeling tendencies are reflected at finer feature resolutions. We have $C$ composed of annotator-dependent kernel subsets $\{C_k\}_{k=1}^{N+1}$, with each kernel subset $C_k$ capturing the $k$-th annotator's labeling tendency. Note that the set size of $C$ is $N+1$ rather than $N$, since we have the last kernel subset $C_{N+1}$ represent the labeling tendency \textit{shared} by the annotators (\eg, background).

In the forwarding process of tendency learning layers, given the input feature map $f$, we select different $C_k$ to perform convolution operation for each pixel, and derive the output feature map $f^\prime \in \mathbbm{R}^{H\times W\times D}$, where $D$ denotes the channel size. To select the proper kernel subset for the forwarding process, our annotator assigner $G$ would produce an annotator mask $M^{A} \in \mathbbm{R}^{H\times W}$, where each element $m^{A}_{u, v} \in M^{A}$ indicates the annotator index for pixel $(u, v)$ (as detailed in Sect.~\ref{subsec:learn_to_assign}). If $m^{A}_{u, v} = k$ (recall that $k$ is the annotator index/label), we then select the $k$-th kernel subset $C_k$ to perform convolution operation for pixel $(u, v)$. Thus, at the tendency learning layer, we have the output feature map calculated as:
\begin{equation}
\begin{aligned}
\mathbf{f^\prime_{u, v}} = conv(\mathbf{f_{u, v}}, C_k \mid m^{A}_{u, v} = k).
\end{aligned}
\end{equation}

To ensure that each kernel subset only captures the labeling tendency of the corresponding annotator, we have $C_k$ only observe data labeled by the exactly $k$-th annotator while $C_{N+1}$ learn from all the annotators. To this end, we calculate the cross-entropy loss between the predicted segmentation mask $\hat{y}_i$ and the ground-truth segmentation mask $y_i$ in an annotator-aware manner. More precisely, given training data $(x_i, y_i, a_i)$ (recall that one input image $x_i$ is annotated by one single annotator $a_i$), we define the \textit{annotator-aware segmentation loss} $\mathcal{L}_{seg}^{(k)}$ for $k \in \{1,2,\dotsb,N\}$:
\begin{equation}
\begin{aligned}
\mathcal{L}_{seg}^{(k)} = \sum_{i} CE(\hat{y}_{i}, y_{i}) \mathbbm{1}_k(a_i = k).
\end{aligned}
\end{equation}
Note that $CE(\cdot, \cdot)$ denotes the cross-entropy loss, and $\mathbbm{1}_k(\cdot)$ is an indicator function that outputs $1$ when the image is annotated by the $k$-th annotator ($0$ otherwise).
Thus, the corresponding kernel subset $C_k$ is updated by:
\begin{equation}
\begin{aligned}
\theta_{C_k}\leftarrow \theta_{C_k} - \frac{\partial \mathcal{L}_{seg}^{(k)}}{\partial \theta_{C_k}}, \quad k \in \{1,2,\dotsb,N\}.
\end{aligned}
\end{equation}
On the other hand, since $C_{N+1}$ represents the kernel subset with tendency \textit{shared} by all annotators, we have the aggregated annotator-aware segmentation loss $\mathcal{L}_{seg}$ as:
\begin{equation}
\begin{aligned}
\mathcal{L}_{seg} = \sum_{k=1}^{N} \mathcal{L}_{seg}^{(k)}
\equiv \sum_{i} CE(\hat{y}_{i}, y_{i}).
\end{aligned}
\end{equation}
Thus, this shared kernel subset $C_{N+1}$ is updated by:
\begin{equation}
\begin{aligned}
\theta_{C_{N+1}}\leftarrow \theta_{C_{N+1}} - \frac{\partial \mathcal{L}_{seg}}{\partial \theta_{C_{N+1}}}.
\end{aligned}
\end{equation}
As for updating the remaining part of the segmentation model $F$ (excluding annotator-dependent kernels $C$), we apply the overall segmentation loss (\ie, $\mathcal{L}_{seg}$) for optimizing the parameters $\theta_{F\setminus C}$:
\begin{equation}
\begin{aligned}
\theta_{F\setminus C}\leftarrow \theta_{F\setminus C} - \frac{\partial \mathcal{L}_{seg}}{\partial \theta_{F\setminus C}}.
\end{aligned}
\end{equation}

With the above learning strategies, each kernel subset $C_k \in C$ would describe the desirable labeling tendency of each annotator, realizing segmentation interpretability at the \textit{annotator} level.

\subsection{Learning to Assign Labeling Tendencies}
\label{subsec:learn_to_assign}

As stated in Sect.~\ref{subsec:learn_to_represent}, learning of annotator-dependent kernel subset $C_k$ is based on the assignment of the annotator mask $M^{A}$. We now describe how we learn the annotator assigner $G$ for predicting this mask $M^{A}$, offering segmentation interpretability at the \textit{assignment} level.

\paragraph{Prototype-Bank Based Annotator Assigner $\boldsymbol{G}$}
As depicted in Figure~\ref{fig:overview}, we have an annotator assigner $G$ take the input image $x_i$ and output the annotator mask $M^{A} \in \mathbbm{R}^{H \times W}$, allowing the subsequent tendency learning layers to learn annotator-dependent kernels. To offer assignment-traceable interpretability, this assigner $G$ is based on the learning of \textit{annotator-dependent prototype bank} $P$, which serves as memory-bank-like~\cite{gong2019memorizing} bases with visually explainable bases/prototypes. Specifically, $P$ contains $N+1$ groups of annotator-dependent prototypes:
\begin{equation}
\begin{aligned}
P = [P_1, P_2, \dotsb, P_{N+1}] \in \mathbbm{R}^{d \times ((N+1) \times Q)},
\end{aligned}
\end{equation}
where prototype group $P_k \in P$ includes $Q$ trainable and explainable prototypes $\mathbf{p}_k \in \mathbbm{R}^d$ describing local image patterns associated with the tendency of the $k$-th annotator:
\begin{equation}
\begin{aligned}
P_k = [\mathbf{p}_k^1, \mathbf{p}_k^2, \dotsb, \mathbf{p}_k^Q] \in \mathbbm{R}^{d \times Q}.
\end{aligned}
\end{equation}
With the learned prototype bank $P$, we can derive $M^{A}$ for $x_i$, reflecting the segmentation tendency for each local regions. To be more precise, we feed $x_i$ into the prototype encoder $E$ to derive the resulting feature map whose size is $h \times w \times d$, with each local feature $\mathbf{E(x_i)_{u, v}} \in \mathbbm{R}^{d}$ indicating the feature of the location $(u, v)$. Then, we calculate the similarity between $\mathbf{E(x_i)_{u, v}}$ and the prototypes across annotators and identify the annotator tendency of interest. That is, for each $\mathbf{E(x_i)_{u, v}}$, we calculate the cosine similarity score $s$ with each prototype $\mathbf{p}_k \in P_k$, and select the highest score $s_k^{\ast}$ for $k \in \{1,2,\dotsb,N+1\}$:
\begin{equation}
\begin{aligned}
s_k^{\ast} &= \mathop{\max}_{\mathbf{p}_k \in P_k} sim(\mathbf{E(x_i)_{u, v}},\: \mathbf{p}_k),
\end{aligned}
\end{equation}
where $sim(\cdot, \cdot)$ denotes the cosine similarity function. With $N+1$ selected scores, we thus have the \textit{soft} version of the annotator mask $M \in \mathbbm{R}^{h \times w \times (N+1)}$, in which each vector $\mathbf{m}_{u, v} \in M$ indicates scores favoring the labeling tendencies across $N+1$ annotators at the location $(u, v)$:
\begin{equation}
\begin{aligned}
\mathbf{m}_{u, v} = [s_1^{\ast}, s_2^{\ast}, \dotsb, s_{N+1}^{\ast}] \in \mathbbm{R}^{N+1}.
\end{aligned}
\end{equation}
To derive the final annotator mask $M^{A} \in \mathbbm{R}^{H \times W}$, we simply apply the argmax operation on $M \in \mathbbm{R}^{h \times w \times (N+1)}$ for selecting the annotator index for each position and up-sample it to match the image resolution size:

\begin{equation}
\begin{aligned}
M^A = UP_{r} (argmax(M)),
\end{aligned}
\end{equation}
where $argmax(\cdot)$ is performed along each vector $\mathbf{m}_{u, v} \in M$, and $UP_{r}(\cdot)$ denotes the up-sampling operation with the ratio of $r = (H/h, W/w)$.

To learn a reasonable $M^{A}$, we then introduce the pseudo annotator mask $M^{pseu} \in \mathbbm{R}^{h \times w \times (N+1)}$ as the guidance, where $M^{pseu}$ assigns a specific annotator for boundary prediction while employing the \textit{shared} annotator to handle the remaining consensus regions (\eg, background). The motivation is that, the boundaries are the most uncertain regions and thus are the major variance of segmentation tendencies \cite{zhang2020disentangling}. To produce $M^{pseu}$, we identify uncertain regions of an input image $x_i$ by pre-training a \textit{vanilla} segmentation model $F_{van}$ (\eg, a plain U-Net~\cite{ronneberger2015u}) on only $\{X, Y\}$ but $A$ (and then freeze for later uses). Given an input $x_i$, the output of $F_{van}$ results in the uncertainty mask $U \in \mathbbm{R}^{H \times W}$, where each element of $U$ indicates if the region is uncertain (represented by $1$) or not (represented by $0$). After that, we assign specific labeling tendencies for those uncertain regions while assigning the shared tendency for the remaining consensus ones. Thus, we would assign the annotator index/label $a_i=k$ if the element is $1$ on $U$, and specify $N+1$ otherwise, and then we convert the annotator indexes to one-hot-encoded probabilities. Finally, we perform down-sampling to obtain the final $M^{pseu}$ whose size is the same as that of $M \in \mathbbm{R}^{h \times w \times (N+1)}$.

\paragraph{Updating Annotator Assigner $\boldsymbol{G}$}
With the pseudo annotator mask $M^{pseu}$ as the guidance, we are able to learn the annotator assigner $G$ using the following \textit{annotator assignment loss} $\mathcal{L}_{assign}$, which computes the cross-entropy loss ($CE$) between $M_{i}$ and $M^{pseu}_{i}$ with respect to $x_i$:
\begin{equation}
\begin{aligned}
\mathcal{L}_{assign} = \sum_{i} CE(softmax(M_{i}), M^{pseu}_{i}).
\end{aligned}
\end{equation}
Note that $softmax(\cdot)$ transforms logits into a probability distribution suitable for the cross-entropy loss calculation. Thus, the optimization of $G$ (only updating $E$ and $P$) can be formulated as follows:
\begin{equation}
\begin{aligned}
\theta_{E, P}\leftarrow \theta_{E, P} - \frac{\partial \mathcal{L}_{assign}}{\partial \theta_{E, P}}.
\end{aligned}
\end{equation}

It can be seen that, the motivation behind the above optimization is to produce prototypes $\mathbf{p}_k \in P_k$ being substantially representative to the uncertain local image regions of those labeled by the $k$-th annotator ($k$ from $1$ to $N$), while prototypes $\mathbf{p}_{N+1} \in P_{N+1}$ are encouraged to describe the remaining local regions (\eg, obvious foreground or background regions). With the completion of training prototype banks $P$ and assigner $G$, our model would offer explainable and traceable segmentation results at the assignment level.

\subsection{Visual Interpretability during Inference}
\label{subsec:inference}
With learned $G$, $F$, and $C$, we now explain how interpretability at both \textit{annotator} and \textit{assignment} levels can be achieved. For the annotator-level explanation, the test input image is fed into $G$ and $F$ for deriving the annotator mask and segmentation output. Thus, given a patch location of interest, we are able to answer the question of ``who" by pinning out the annotator index on the annotator mask.

As for the assignment-level explanation, we are able to visualize the most similar patches (via prototype identification with the patch size of $\frac{H}{h} \times \frac{W}{w}$) for each pixel and answer the question of ``why". That is, our model assigns the annotator for predicting the label for that pixel, since the input patch is visually similar to those annotated by the corresponding annotator during training.
\section{Experiments}
\label{sec:experiments}

\subsection{Datasets and Implementation Details}

\subsubsection{Datasets}
\label{subsubsec:datasets}
\textbf{PASCAL VOC 2012}~\cite{everingham2010pascal} (referred to below as \textbf{PASCAL VOC}) contains 20 foreground classes and one background class in realistic scenes. We adopt commonly used augmented version~\cite{hariharan2011semantic} with 10,582/1,449/1,456 images for training/validation/testing splits. We follow~\cite{chen2018encoder} and report the mean intersection-over-union (mIoU) across the 21 classes on the validation split.

\textbf{CVC-EndoSceneStill}~\cite{vazquez2017benchmark} (referred to below as \textbf{EndoScene}) is an endoluminal scene segmentation benchmark for polyps from colonoscopy images, and contains foreground (polyps) and background classes. The dataset is divided into 547/182/182 images for the training/validation/testing splits. We report the mIoU across the two classes on the testing split.

\textbf{LIDC-IDRI}~\cite{armato2011lung} holds 1018 lung computed tomography (CT) scans, resulting in 8,882/1,996/1,992 images as the training/validation/testing splits. Each image is annotated by four thoracic radiologists, and we leave one label per image for training data by following \cite{zhang2020disentangling, baumgartner2019phiseg}, and we use single-label ground-truth for evaluation~\cite{zhang2020disentangling}. We report the mIoU, mean-DICE (shortened as m-DICE) and DICE (considering only the foreground class) on the testing split.

\begin{figure}[t!]
	\centering
	\includegraphics[width=0.47\textwidth]{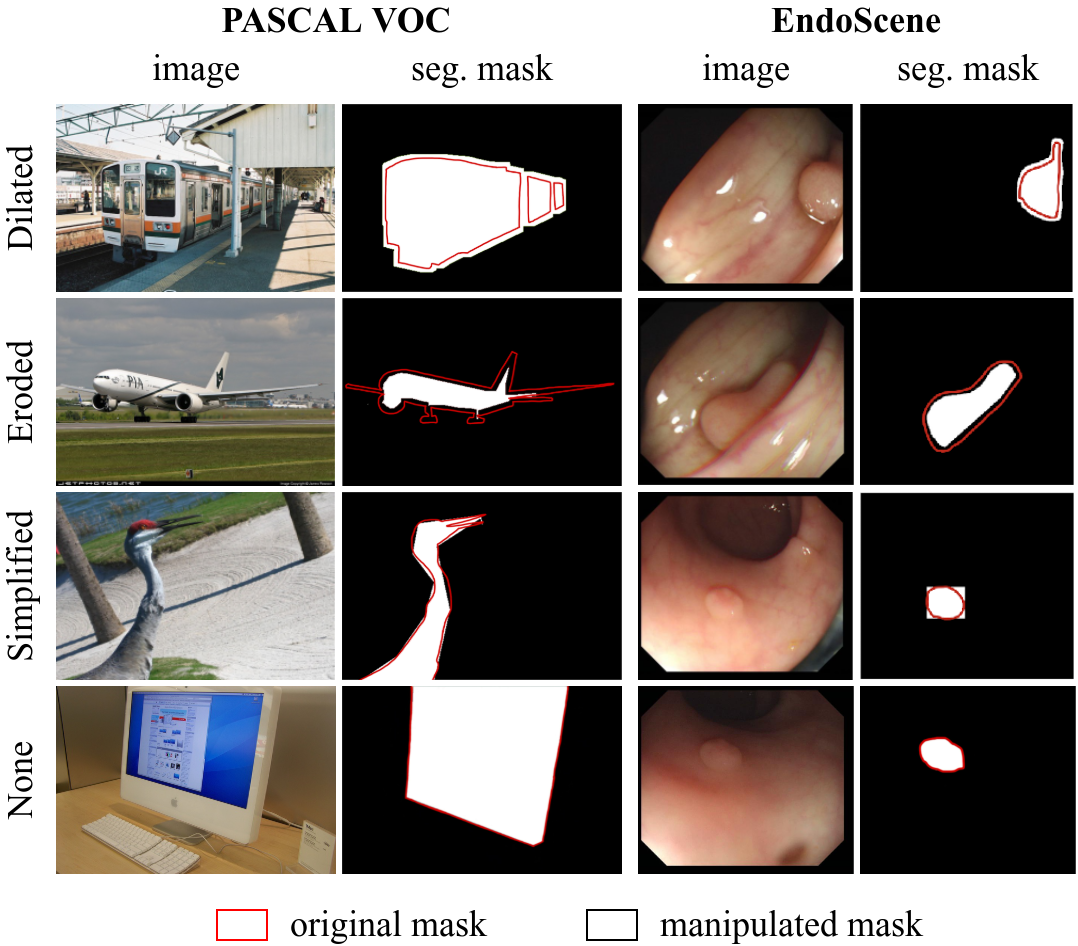}
 	\vspace{0mm}
    \caption{\textbf{Examples of manipulated segmentation masks for imitating tendencies of multi-annotators.}
    Note that the foreground is shown in white for simplicity.}
    \vspace{0mm}
	\label{fig:syn_example}
\end{figure}

Since only the ground-truth labels of \textbf{LIDC-IDRI} are annotated by multiple (four) experts, we explain how we manipulate multi-annotator labels for the first two datasets.
For PASCAL VOC and EndoScene, we partition the training images into four subsets (each stands for one annotator), and we manipulate their original masks with pre-defined labeling tendencies via four morphological operations: \textit{dilated}, \textit{eroded}, \textit{simplified}, and \textit{none}, respectively. As shown in Figure~\ref{fig:syn_example}, \textit{dilated} tends to over-segment, while \textit{eroded} tends to under-segment the boundaries. \textit{Simplified} shows more straight contours, and \textit{none} indicates the original masks.

\subsubsection{Implementation Details}

We implement our proposed tendency-preserving segmentation model $F$ with two alternative architectures, U-Net~\cite{ronneberger2015u} and DeepLabv3+~\cite{chen2018encoder} (shortened as DLv3+). For each architecture, we employ two encoder backbones pretrained on ImageNet~\cite{deng2009imagenet}, ResNet101~\cite{he2016deep} (shortened as Res101) and EfficientNet-B5~\cite{tan2019efficientnet} (shortened as Eff-B5). Prototype encoder $E$ and the vanilla segmentation model $F_{van}$ are implemented simply with the U-Net~\cite{ronneberger2015u} architecture with ResNet50~\cite{he2016deep} as the encoder backbone pretrained on ImageNet~\cite{deng2009imagenet}.

\begin{figure*}[t!]
	\centering
\includegraphics[width=0.94\textwidth]{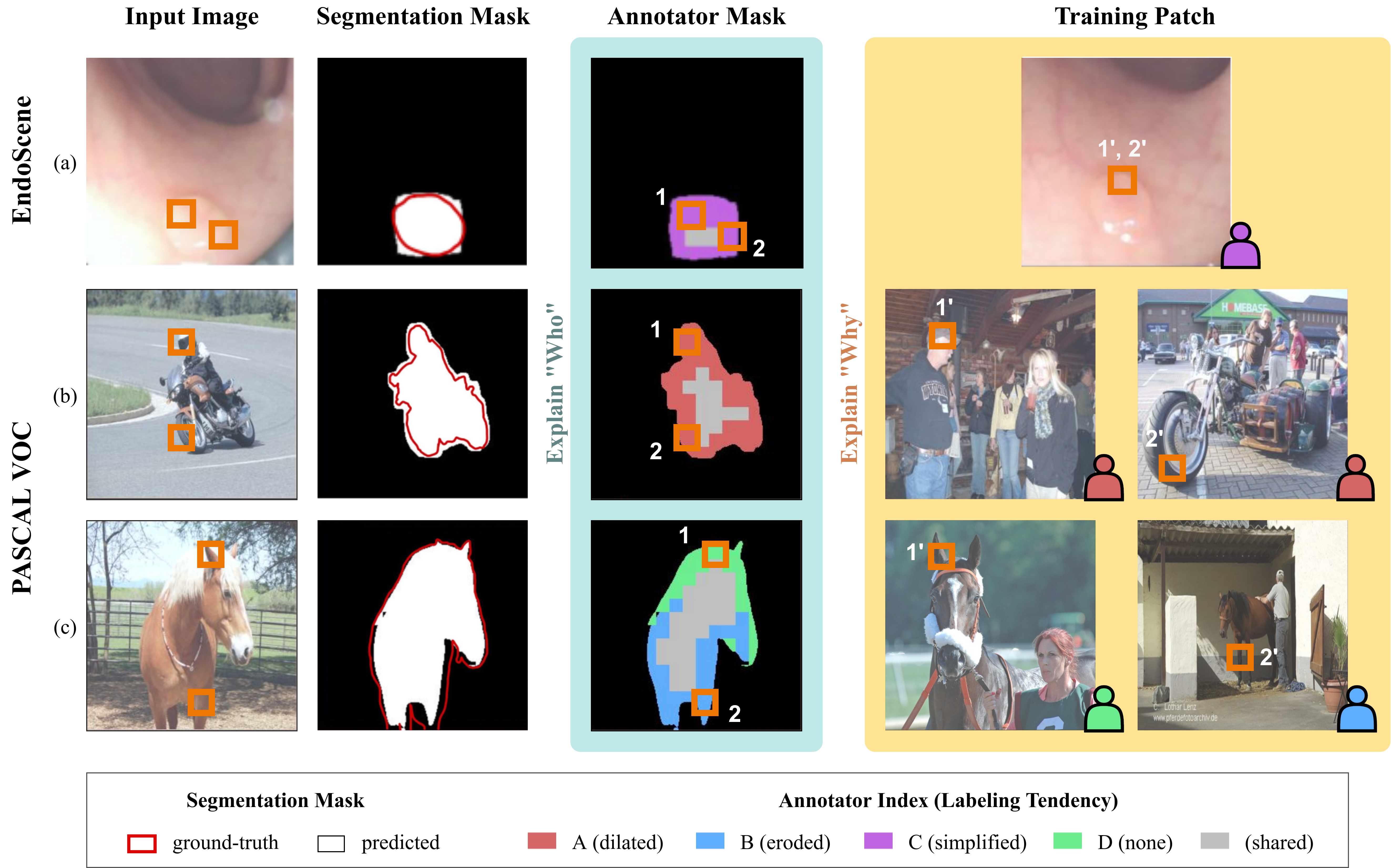}
 	\vspace{0mm}
    \caption{\textbf{Case studies for interpretability on EndoScene and PASCAL VOC.} Each row presents an input image, segmentation prediction, and its interpretability at annotator/assignment levels (\ie, explaining who/why). Note that the original ground-truth masks are in red. Best viewed in color.}
    \vspace{0mm}
	\label{fig:casestudy_syn}
\end{figure*}

\subsection{Case Studies for Interpretability}
\label{subsec:case_study}
To verify the interpretability of our TAX, in Figure~\ref{fig:casestudy_syn} we study annotator (who) and assignment (why) explanations on EndoScene, PASCAL VOC, and LIDC-IDRI.

\paragraph{EndoScene}
As the first case shown in Figure~\ref{fig:casestudy_syn}(a), we see that our model produced a simplified and rectangle-like segmentation mask, compared to the original, non-manipulated ground-truth mask in red. One would raise the question about whose labeling tendency results in such a segmentation output. After examination, we notice that the predicted segmentation mask for this input image was produced by the annotator with \textit{simplified} labeling tendency (as defined in Sect.~\ref{subsubsec:datasets}). The annotator-specific results (\ie, annotator masks) were as depicted in purple, reflecting his/her labeling tendency of such annotation. Following such interpretability at the annotator level, the subsequent question one would dig into is why the model chooses to utilize the labeling tendency of this annotator. To answer this question, our annotator assigner $G$ would be applied to identify the prototypes specifically selected for performing such segmentation (see Sect.~\ref{subsec:learn_to_assign}), suggesting the training patches with high similarity to such input patches. This can be visually verified by comparing the input and training patches in orange bounding boxes (denoted with 1$^\prime$, 2$^\prime$), depicted in the first and last images/columns in Figure~\ref{fig:casestudy_syn}(a), respectively. Since EndoScene contains only polyps images, the selected training patches do not exhibit significant visual appearance variety (\ie, limited assignment-level interpretability due to data). To better verify the identifying ability of the annotator assigner, we then perform the case studies on PASCAL VOC below.

\paragraph{PASCAL VOC}
We now study a more challenging multi-class scenario on PASCAL VOC. In Figure~\ref{fig:casestudy_syn}(b), the second case study shows that our model predicted an over-segmented result compared to the original, non-manipulated ground-truth segmentation mask. The predicted annotator mask (in red) can be seen to be consistent with the annotator with prior \textit{dilated} labeling tendency (as defined in Sect.~\ref{subsubsec:datasets}). Take the annotator mask around the helmet region for example. Our annotator assigner identified the most similar training patch (a head with a cap) via the learned annotator assigner and its prototype bank. Also, for the motorcycle tire in the input image, its segmentation mask corresponds to the same \textit{dilated} labeling tendency, with the most similar training patch shown in the last image of Figure~\ref{fig:casestudy_syn}(b).

\begin{figure}[t!]
	\centering
\includegraphics[width=0.47\textwidth]{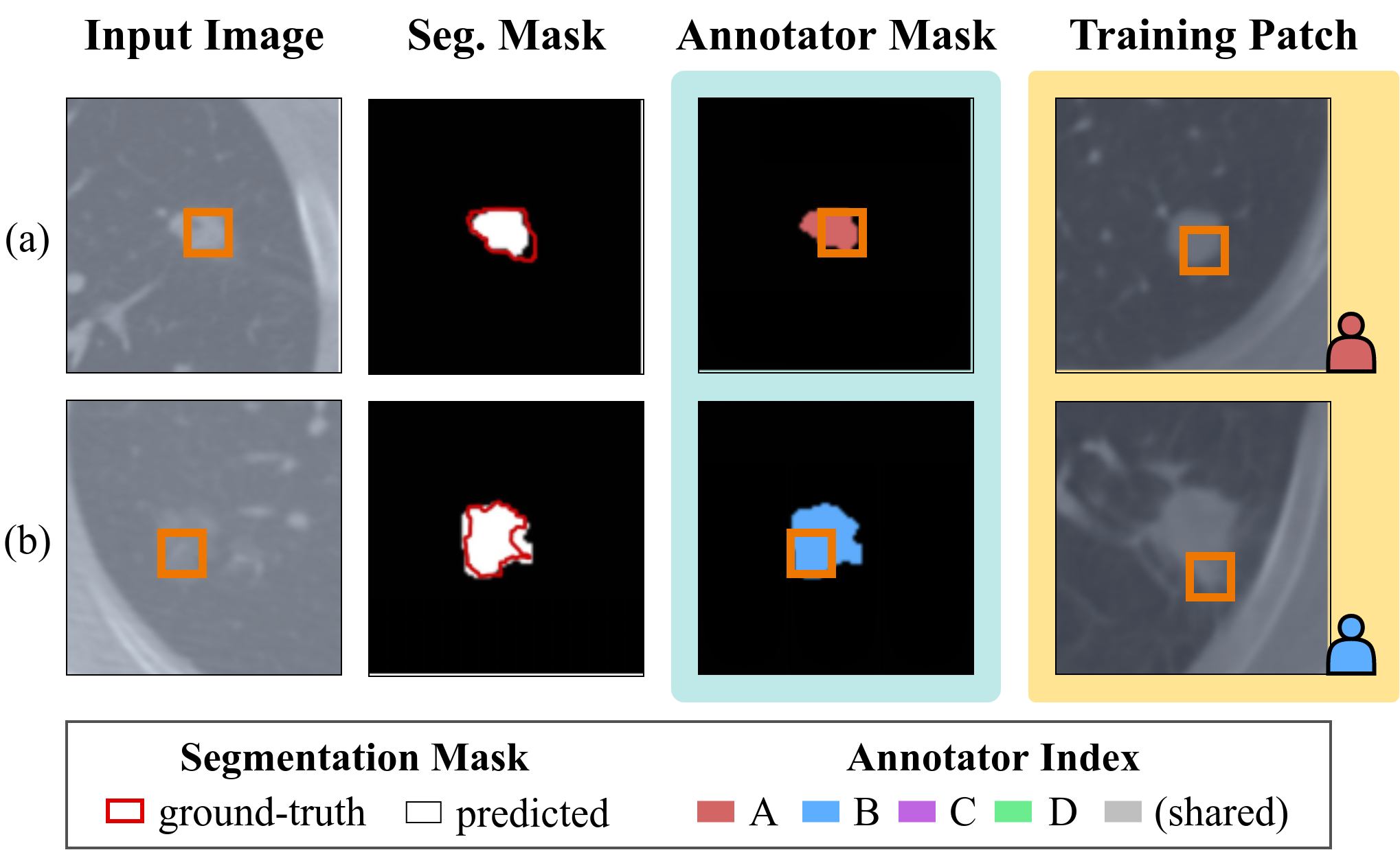}
 	\vspace{0mm}
    \caption{\textbf{Case studies for interpretability on LIDC-IDRI.} Note that the ground-truth masks are in red. Best viewed in color.}
    \vspace{0mm}
	\label{fig:casestudy_real}
\end{figure}

In the third case shown in Figure~\ref{fig:casestudy_syn}(c), the predicted segmentation exhibited hybrid segmentation tendencies on the boundary, where the head of the horse was segmented with fitting boundaries, while its feet were under-segmented. For the annotator mask shown in Figure~\ref{fig:casestudy_syn}(c), we notice that these two parts were indeed associated with different labeling tendencies, \ie, \textit{none} and \textit{eroded}. As for tracing back the most similar training patches, the identified patches are shown in the last two images of Figure~\ref{fig:casestudy_syn}(c), confirming the visual and semantic similarities to the input ones. This case study confirms that, our model can delicately assign different annotators for local segmentation prediction. From the above case studies, we verify that our model not only performs segmentation with interpretability at the annotator level, the segmentation mask and its associated patch can be traced back to the training ones with the same labeling tendency, realizing interpretability at the assignment level.

\paragraph{LIDC-IDRI}
We now perform case studies on the real-world multi-annotator dataset of LIDC-IDRI. As shown in Figure~\ref{fig:casestudy_real}, the predicted segmentation masks of the first and second cases can be recognized as distinct tendencies, with annotator masks shown in red and blue, respectively. Again, our model is able to explain such segmentation tendency by tracing back to the most similar patches in the training data. It is worth repeating that, prior works like~\cite{zhang2020disentangling, baumgartner2019phiseg, kohl2018probabilistic} are not designed to offer the above interpretability, and are designed to predict a unified segmentation output without explanation of labeling tendency or training data similarity. Thus, the use of our model would be desirable especially for real-world applications such as medical image analysis.

\subsection{Quantitative Analyses}
\subsubsection{Segmentation Performance}
We now offer quantitative results to verify the effectiveness of our approach using PASCAL VOC, EndoScene, and LIDC-IDRI. As noted in Sect.~\ref{subsec:learn_to_represent}, our approach is applicable to different segmentation architectures (\eg, U-Net and DeepLabv3+), and thus we present segmentation results using different backbones in Table~\ref{table:comparison_seg}. Particularly, \textit{baseline} models in Table~\ref{table:comparison_seg} denote non-explainable vanilla segmentation models simply learned from the entire training dataset with manipulated ground-truth labels. From Table~\ref{table:comparison_seg}, we see that we achieved comparable mIoU as baseline models did, suggesting that deploying our TAX learning strategy would not significantly affect the segmentation performance, while interpretability can be additionally offered.

Furthermore, we conduct experiments on LIDC-IDRI for multi-annotator segmentation. We compare with state-of-the-art methods of~\cite{kohl2018probabilistic, baumgartner2019phiseg, asman2012formulating, zhang2020disentangling}, which are designed for this setting yet not able to offer output interpretability. As shown in Table~\ref{table:comparison_sota}, our model achieved comparable segmentation performances as the competitors did. It is worth noting that, we qualitatively and quantitatively verify the interpretability of our model in Sections~\ref{subsec:case_study} and~\ref{subsec:annotator_assignment_explanation}, while the above methods were not able to provide such explanations. This again confirms the effectiveness of our method for real-world semantic segmentation.
% It is worth repeating that, thanks to our TAX learning framework, our model is able to provide annotator and assignment-level explanations (as discussed in Sect.~\ref{subsec:case_study}) which other competitors cannot reach.

\begin{table}[!tp]
\resizebox{0.47\textwidth}{!}{
\begin{tabular}{c|c|c|c|c|c|c|c}
\hline
\multirow{2}{*}{Archit.} & \multirow{2}{*}{Backbone} & \multicolumn{2}{c|}{PASCAL VOC} & \multicolumn{2}{c|}{EndoScene} & \multicolumn{2}{c}{LIDC-IDRI}     \\ \cline{3-8} 
                              &                           & Baseline       & Ours           & Baseline         & Ours           & Baseline       & Ours         \\ \hline
\multirow{2}{*}{U-Net}        & Res101                 & 69.28          & \textbf{69.31} & \textbf{82.80}   & 82.60          & 54.02          & \textbf{54.52}          \\ \cline{2-8} 
                              & Eff-B5                    & 73.24          & \textbf{73.95} & \textbf{82.51}   & 82.50          & 54.55          & \textbf{55.27}\\ \hline
\multirow{2}{*}{DLv3+}   & Res101                 & \textbf{74.86} & 74.72          & 81.90            & \textbf{82.80} & \textbf{55.20}          & 54.93\\ \cline{2-8} 
                              & Eff-B5                    & 77.07          & \textbf{77.15} & 83.15            & \textbf{83.25} & 54.32          & \textbf{55.30} \\ \hline
\end{tabular}}
\vspace{3mm}
\centering
\caption{\textbf{Semantic segmentation performances on PASCAL VOC, EndoScene and LIDC-IDRI in terms of mIoU (\%).} Note that our learning method can be applied to different backbones and produce comparable performances, while additional interpretability can be offered (as verified in Table~\ref{table:comparison_tend}).}
\label{table:comparison_seg}
\end{table}
\begin{table}[!tp]
\resizebox{0.47\textwidth}{!}{
\begin{tabular}{c|cccc|c}
\hline
\multirow{2}{*}{Method} & P. U-Net & PHiSeg & S. STAPLE & Zhang \etal & \multirow{2}{*}{Ours} \\
& \cite{kohl2018probabilistic} & \cite{baumgartner2019phiseg} & \cite{asman2012formulating} & \cite{zhang2020disentangling} & \\
\hline
DICE & 52.38 & \textbf{54.08} & - & - & 53.25 \\
\hline
m-DICE & - & - & 62.35 & 68.12 & \textbf{68.25} \\
\hline
\end{tabular}}
\vspace{3mm}
\centering
\caption{\textbf{Semantic segmentation on LIDC-IDRI in terms of DICE and m-DICE (\%).} While our TAX is observed to achieve comparable results, the state-of-the-arts for multi-annotator segmentation are \textit{not} able to provide output explanation as ours does.
% while our model can achieve satisfactory performances while additional interpretability can be provided (as verified in Table~\ref{table:comparison_tend})
}
\label{table:comparison_sota}
\end{table}

\subsubsection{Annotator and Assignment-Level Explanations}
\label{subsec:annotator_assignment_explanation}
Earlier in Sect.~\ref{subsec:case_study}, we qualitatively demonstrated that our method is capable of offering visual explanations at the annotator and assignment levels. Here, we assess the explanation abilities in quantitative manners.

To quantify the effectiveness of our annotator-level explanations, we examine whether each individual convolution kernel subset $C_k$ exactly learns the labeling tendency of the corresponding annotator $k$. In other words, we choose to assess the segmentation performance of the learned tendency-preserving segmentation model $F$ using only $C_k$. Specifically, for each test image $x_t$ with ground truth $y_t$ annotated by a particular labeling tendency (\ie, labeled by the annotator $k$), we then calculate the mIoU between the predicted $\hat{y}_t$ and $y_t$. As presented in Table~\ref{table:comparison_tend}, our method achieved superior performances by significant margins over baseline models (\ie, vanilla segmentation models) which are not designed to handle individual labeling tendency. This thus verifies that our tendency learning layer succeeds in describing the labeling tendency of each annotator, supporting our interpretability at the annotator level.

\begin{table}[!tp]
\resizebox{0.47\textwidth}{!}{
\begin{tabular}{c|c|c|c|c|c|c|c}
\hline
\multirow{2}{*}{Archit.} & \multirow{2}{*}{Backbone} & \multicolumn{2}{c|}{PASCAL VOC} & \multicolumn{2}{c|}{EndoScene} & \multicolumn{2}{c}{LIDC-IDRI}     \\ \cline{3-8} 
                              &                        & Baseline       & Ours           & Baseline     & Ours           & Baseline       & Ours  \\ \hline
\multirow{2}{*}{U-Net}        & Res101                 & 62.60          & \textbf{66.01} & 77.84        & \textbf{80.00} & 53.20          & \textbf{53.42} \\ \cline{2-8} 
                              & Eff-B5                 & 65.47          & \textbf{70.56} & 77.72        & \textbf{79.55} & 52.85          & \textbf{54.20} \\ \hline
\multirow{2}{*}{DLv3+}        & Res101                 & 67.23          & \textbf{69.15} & 77.53        & \textbf{80.00} & \textbf{53.85} & 53.70 \\ \cline{2-8} 
                              & Eff-B5                 & 69.12          & \textbf{72.78} & 78.32        & \textbf{81.16} & 52.55          & \textbf{54.30} \\ \hline
\end{tabular}}
\vspace{3mm}
\centering
\caption{\textbf{Evaluation of segmentation with labeling tendency on PASCAL VOC, EndoScene and LIDC-IDRI in mIoU (\%).} Note that, for PASCAL VOC and EndoScene, pixel-level labels are manipulated by instructions presented in Sect.~\ref{subsubsec:datasets}, while LIDC-IDRI contains labels from multi-annotators.}
\label{table:comparison_tend}
\end{table}

As for the assessment of assignment-level explanations, we choose to evaluate whether our learned annotator assigner $G$ is able to identify the visual similarity between prototypes and input images. To conduct this quantitative evaluation, we again have each test image $x_t$ with ground-truth labels annotated by a particular labeling tendency (\eg, $a_t = k$, one of the morphological operations). We then feed $x_t$ into $G$ to derive the annotator mask $M^A$, followed by a majority vote across pixels $(u, v)$ in $M^A$ for calculating the assignment accuracy. A considerably high assignment accuracy of \textbf{85.80\%} was obtained on PASCAL VOC, confirming that our annotator assigner is able to select proper prototypes visually similar to the input patches. This thus further supports the interpretability of our model at the assignment level.
\section{Conclusion}
\label{sec:conclusion}

In this paper, we aim at introducing interpretability to segmentation models in the multi-annotator scenario, offering explanations to the segmentation output at both annotator (who) and assignment (why) levels. With our proposed framework of TAX, a proper set of convolution kernels would be derived for describing the labeling tendency of each individual annotator, while the learned prototype bank allows one to trace back the segmentation outputs to the training image data. Our case studies on manipulated and real-world datasets qualitatively supported the interpretability of our proposed method. Moreover, our quantitative experiments verified the use of TAX with state-of-the-art network models for producing satisfactory performances.

%%%%%%%%% REFERENCES
{\small
\bibliographystyle{ieee_fullname}
\bibliography{egbib}
}

\end{document}